\pdfoutput=1

\documentclass[11pt]{article}

\usepackage{acl}

\usepackage{times}
\usepackage{latexsym}
\usepackage{amsmath}
\usepackage{url}

\usepackage[T1]{fontenc}

\usepackage[utf8]{inputenc}

\usepackage{microtype}

\usepackage{adjustbox}
\usepackage{graphicx}
\usepackage{multirow}
\usepackage{booktabs}
\usepackage{todonotes}
\usepackage{float}
\usepackage{xstring}
\usepackage{enumitem}
\usepackage[bottom]{footmisc}

\definecolor{lightblue}{HTML}{E0ECF7}
\definecolor{darkblue}{HTML}{092E6B}

\title{Learning from data in the mixed adversarial non-adversarial case:\\ {\em Finding the helpers and ignoring the trolls}}
\author{Da Ju\\
  Meta AI \\
  \\\And
  Jing Xu\\
  Meta AI \\
  \\\AND
  Y-Lan Boureau\\
  Meta AI \\
  \\\And
  Jason Weston \\
  Meta AI \\
}

\begin{document}
\maketitle
\begin{abstract}
The promise of interaction between intelligent conversational agents and humans is that models can learn from such feedback in order to improve.  
Unfortunately, such exchanges in the wild will not always involve human utterances that are benign or of high quality, and will include a mixture of engaged (helpers) and unengaged or even malicious users (trolls). In this work we study how to perform robust learning in such an environment.
We introduce a benchmark evaluation, SafetyMix, 
which can evaluate methods that learn  safe vs. toxic language in a variety of adversarial settings to test their robustness.
We propose and analyse several mitigating learning algorithms
that identify trolls either at the example or at the user level.
Our main finding is that user-based methods, that take into account that troll users will exhibit adversarial behavior across multiple examples,  
work best in a variety of settings on our benchmark.
We then test these methods in a further real-life setting of conversations collected during  deployment, with similar results.
\end{abstract}

\section{Introduction}

Humans learn through interactions with other humans, while simultaneously learning who to trust and who not to trust \cite{subrahmanian2021detecting}. 
When models interact with humans in natural situations, one might expect similar challenges.
In human-bot conversations the problem of adversarial interactions can be exacerbated because it is known that certain groups of humans can behave poorly towards bots in real life deployments \cite{chatbot_curse,wolf2017we}, where we refer to such humans as  ``trolls'' \cite{tomaiuolo2020survey,shachaf2010beyond}.
In this work we study automatic methods for models to learn from human interactions, where the goal is to gain maximum learning efficiency from high quality data, while simultaneously being maximally robust to low quality and adversarial data.

Compared to much of the literature on learning with noisy inputs 
\cite{song2020learning} two distinguishing factors of this setting  are that undesirable data from deployment is dependent on the user, and that noise is not only random but can also be adversarial.
We thus construct a benchmark, SafetyMix, particularly for this setting in order to evaluate a variety of methods. We consider different patterns of troll behavior: master trolls that test the limits with difficult inputs, safe (or unsafe)  trolls that label all messages as safe (unsafe), or gaslight trolls that only provide unsafe messages, amongst other variants. This allows us to test which methods work best and under what circumstances they fail.
As well as evaluating standard  robust learning methods that operate per-example/utterance, we also propose three methods that try to detect trust at the user level: per-user removal,  per-user+example removal, and soft per-user robust removal, which all have different characteristics.

\begin{figure}[t!]
  \centering
    \includegraphics[width=0.48\textwidth]{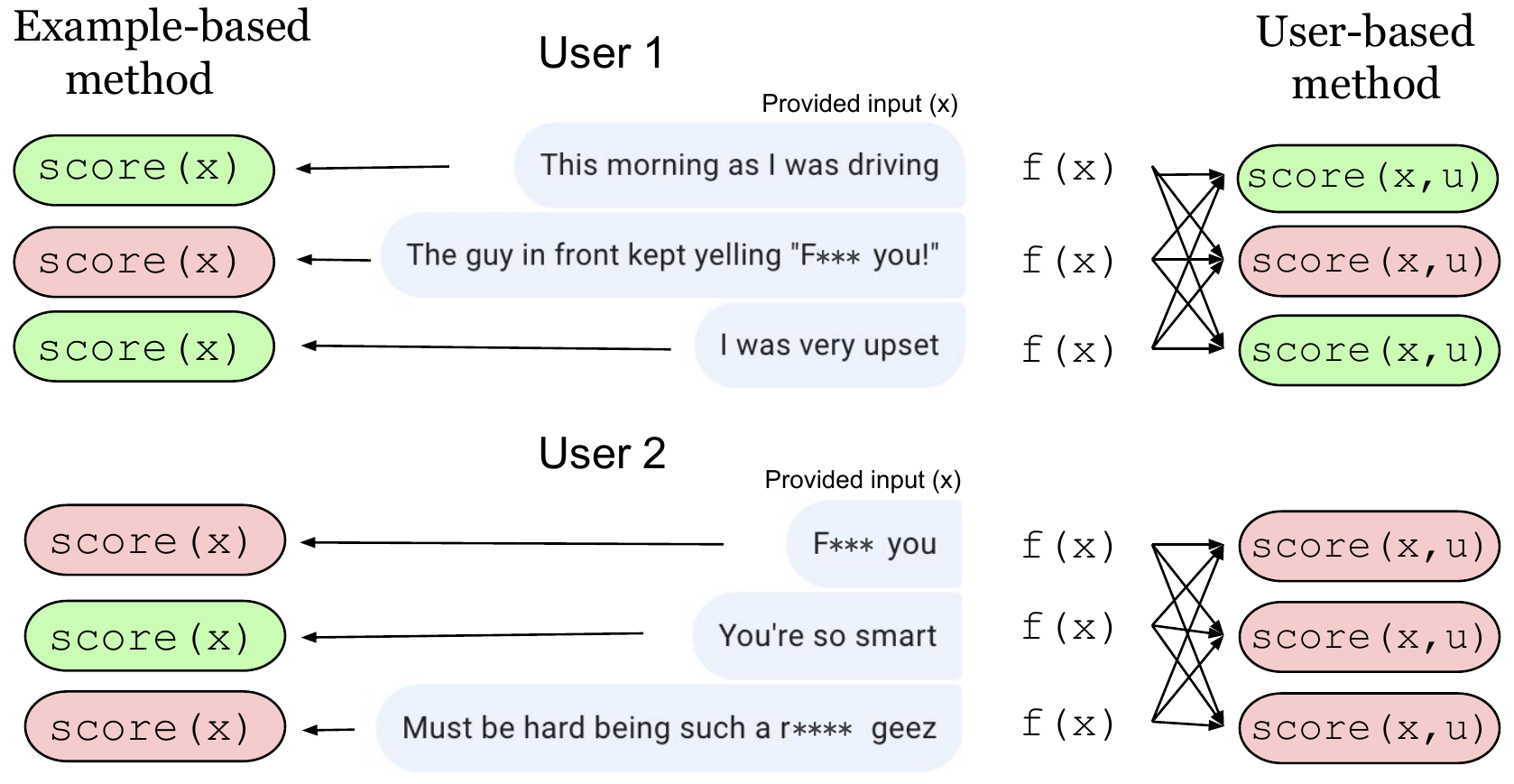}
  \caption{{\bf Detecting Trolls with Example-based vs. User-based methods (Warning: offensive language).} User 1 (helper) provides mostly benign inputs, while User 2's inputs (troll) can be more easily identified as toxic by taking into account scores from all their examples jointly (via a user-based method, right).
    \label{fig:diagram}
    }
\end{figure}

Our main finding is that user-based learning methods that remove  low-quality or malicious feedback by taking into account all the user's behavior perform best on the benchmark in a variety of settings, outperforming per-example based methods. 
User-based methods take into account that troll users with poor behavior tend to be repeat offenders, and this can be spotted by algorithms that, while making decisions at the utterance level, take into account the overall user behavior, i.e. their behavior across other utterances. These results are further verified on a second dataset of conversations collected from the BlenderBot 3 model chatbot deployment \cite{shuster2022bb3}. By verifying results with crowdworkers, we show user-based methods improve troll detection in this real-life case.

While we provide some promising results from some of the proposed algorithms our benchmark SafetyMix also identifies a number of failures, particularly for certain kinds of trolls such as gaslight trolls. Therefore, we expect further improvements to come with advances in 
learning techniques beyond the ones proposed here.
We thus make the code of our experiments and our benchmark SafetyMix publicly available
for further research\footnote{\small{\url{https://parl.ai/projects/trollhunting}}}.

\section{Related Work}

Human-model interaction data for machine learning, and for dialogue research in particular, is commonly collected via expert annotators or crowdworkers \cite{serban2015survey}. While careful instructions \cite{huynh2021survey} can result in good quality feedback or labels to learn from, collection both involves significant monetary costs -- where annotators should be paid well above minimum wage -- and the pool of workers may be limited \cite{moss2020demographic}. An alternative approach is to deploy a system publicly, and collect feedback from organic user interactions. The promise of this approach is that the distribution of data will more closely match those organic users' desires, rather than decided by the researchers themselves when creating datasets \cite{gabriel2020further,roller2020open,shuster2020deploying,ouyang2022training}. Further, a continual deployment of such a system can then potentially keep improving over time \cite{carlson2010toward,kiela2021dynabench,agichtein2006improving,liu2021lifelong,madotto2020continual,shuster2020deploying}, 
constituting what \citet{hancock2019learning} call a ``self-feeding chatbot.''

Unfortunately, a public deployment is likely to have a mixture of varying quality feedback --  high quality engaged users, low quality unengaged users, as well as engaged but deliberately malicious users. The latter includes the use of toxic language \cite{chatbot_curse} as well as deliberately trying to teach the chatbot poor behavior \cite{wolf2017we}.
While much work has centered on detecting  
undesirable behavior
such as toxic language \cite{xu2020recipes,dinan2021anticipating},
trolling \cite{tomaiuolo2020survey,mihaylov2019hunting} 
and bias \cite{dinan2019queens}, less work has studied robust learning from organic conversations with potentially adversarial feedback. 

Learning in the mixed adversarial non-adversarial case is related to
learning from data corrupted with noise, 
a well studied area in machine 
learning. A recent review of this field \cite{song2020learning} characterizes proposed solutions into 
four categories:
robust architectures (e.g., \cite{sukhbaatar2014training,xiao2015learning}), 
robust loss functions (e.g., \cite{ghosh2017robust,liu2020peer}), 
robust regularization  (e.g., \cite{jenni2018deep,goodfellow2014explaining})
and sample selection methods (e.g., \cite{malach2017decoupling,shen2019learning}).
Earlier work goes back to data cleaning \cite{rahm2000data,chapelle2004support,chu2016data}, surrogate losses \cite{natarajan2013learning} and
probabilistic methods \cite{rebbapragada2007class}.
Much of this work focuses on label noise that is either independent or  a function of  the input features, and is typically focused on random noise, mislabeled data and outliers
(see \cite{natarajan2013learning}, Sec. B), and does not model the annotators specifically. The work of  \cite{tanno2019learning} is an example which does model annotator noise.
Still, the setting of labels provided by malicious users is generally not focused on.




\begin{table*}[t!]
\center
\begin{tabular}{lllll}
User Type     &  Input Difficulty & Input Classes  &  &Label Type \\
\hline  
{\bf Helper}        &  Standard         & Both (Safe+Unsafe) &  & Correctly labeled \\
{\bf Troll}         &  Standard         & Both (Safe+Unsafe) &  &   N\% Flipped \\
\hline 
{\bf Master Troll}  & Adversarial       & Both (Safe+Unsafe)  & &  N\% Flipped \\
{\bf Lazy Troll}    & Standard          & Both (Safe+Unsafe)   &  & N\% Noisy \\
{\bf Safe Troll}   & Mixed (Standard+Adversarial)   & Both (Safe+Unsafe)    & & Always marked Safe \\
{\bf Unsafe Troll}    & Mixed (Standard+Adversarial)   & Both (Safe+Unsafe)   &  & Always marked Unsafe \\
{\bf Gaslight Troll}  & Adversarial   & Unsafe only     && Always marked Safe \\
\end{tabular}
\caption{{\bf Helper and Troll Models} used in our {SafetyMix} task experiments. In our experiments troll classes (e.g., Lazy Trolls) are mixed in with Helpers using a 50/50 split in the training data.
\label{table:showtrolls}
}
\end{table*}


\section{The SafetyMix Benchmark} \label{sec:safetymix}

To study the problem of helper and troll identification, we first
construct a benchmark for which we can analyse the properties of various algorithms and measure their success.

\subsection{Modeling  Helpers and Trolls} \label{sec:troll_types}

We consider the case of binary user feedback (safe, unsafe), with $N$ users  providing both inputs and labels. We consider the users can be grouped into a small number of groups $J$ (for example a helper distribution, and a troll distribution). 
For their inputs we consider text sequences of four kinds: standard safe data, standard unsafe data, and adversarially safe and unsafe data. The latter are intended to be difficult for a model to spot and understand, e.g. unsafe text that does not contain any profanity words but can be understood to be unsafe only through its deeper semantic meaning, see \citet{dinan2019build}. In terms of labels, we a model each group  with a probability transition matrix from the true distribution to their individual distribution:
\[
P_i = 
\begin{bmatrix}
P_{00} ~P_{10} \\
P_{01} ~P_{11} \\
\end{bmatrix}
\]
where class 0 means safe, and class 1 is unsafe. Each of the groups has a ratio $G_i$ that represents the fraction of total users that come from this  group  $\sum_i^J G_i = 1$. A {\em helper} would label data correctly, hence having a transition matrix 
$\begin{bmatrix} 
1  ~0 \\ 
0  ~1 \\ 
\end{bmatrix}$
while there are several kind of troll models, e.g. such as  trolls that flip all labels 
$\begin{bmatrix}
0~  1 \\
1~  0 \\
\end{bmatrix}$.

We first consider two main user models in our experiments:
\begin{itemize}
\item{\bf{Helpers:}}  Tend to use standard language, and provide correct feedback.
\item{\bf{Trolls:}} Tend to use standard language, but provide adversarial (flipped) labels. We call 
them a {\bf\em part-time troll} if
they mislabel data only some \% of the time, which is the typical case. 
In our experiments we test different mislabeling rates, while our main results report numbers with N=80\% (we also use this rate for other troll types described below, where applicable). 
\end{itemize}

We also investigate some specific user behavior models in  detailed experiments:

\begin{itemize}
\item{\bf{Master Trolls:}} Test the limits of the model with difficult (adversarial) inputs,  but provide incorrect (flipped) feedback some \% of the time. 
\item{\bf Lazy Troll:} adds noise to the labeling procedure (rather than flipping the label to the incorrect label), which can be seen as mislabeling perhaps due to laziness or mistakes.
\item{\bf Safe Troll:} labels everything as safe, regardless of the input.
\item{\bf Unsafe Troll:} labels everything as unsafe, regardless of the input.
\item{\bf Gaslight Troll:} only provides unsafe and adversarial language,  mislabeled as safe.
\end{itemize}

We summarize these user models in \autoref{table:showtrolls}.
In any given full set of users, one can combine these user models to have a mixture of helpers and trolls.

\subsection{Crowdworker Data and Evaluation}


To build our main SafetyMix benchmark we make use of clean crowdworker data, 
and introduce 
synthetic troll noise in a controlled way using the methods detailed in the previous 
section. This allows us to test different noise models and evaluate the results.
We will address using real-world ``in-the-wild'' data in the next section 
(\autoref{sec:deploy_data}).

We use the conversational safety data collected in \citet{dinan2019build}, 
which is a pool of 30,000 utterances, half of which is collected as standard inputs, and half where crowdworkers were asked to give difficult  adversarial inputs.
10\% of the data is labeled as unsafe, and the rest as safe\footnote{The data is available at: \small{\url{https://parl.ai/projects/dialogue_safety/}}}.
We use this pool of data to construct all the troll types described in
\autoref{table:showtrolls},  assigning these samples to users, where the number of utterances per user are sampled from a normal distribution centered at 10, with a standard deviation of 2.
For any given experiment, we thus sample 200 utterances, and a further 24 for validation purposes while training.
These sizes were chosen to be small so that the otherwise relative simple 
safety classification task was suitably difficult.
We introduce noise in this otherwise clean dataset depending on the troll setting (noise model) being tested.
We reserve the separate valid set of 1000 examples (standard, round 1) from the original paper, which contains 100 unsafe and 900 safe examples for reporting evaluation numbers.

The task is to train a classifier  on the training set, which contains potentially noisy data, that generalizes as well as possible to the clean evaluation set, where we report balanced accuracy. To do this, we evaluate algorithms that identify noisy troll utterances, and filter them from training. For those models we also report the precision and recall of troll utterance detection.

\subsection{Real Deployment Data} \label{sec:deploy_data}

{SafetyMix} is a useful benchmark to compare troll detection methods because we can construct and analyse different setups with known annotations from crowdworkers. 
However, we also need to test our methods in more real-world ``in the wild'' scenarios. For this reason we also use a set of contexts coming 
from the public deployment of BlenderBot 3, a conversational agent that converses with members of the public \cite{shuster2022bb3}.

We take a subset of this data, and consider conversational turns by the human speakers. According to crowdworkers,  31\% of human utterances from this dataset are
deemed poor quality (off topic / ignoring partner, nonsensical / incorrect, rude / inappropriate, or other reason). Moreover, of these utterances 42\% are deemed rude or inappropriate. Hence it appears there may be a significant amount of trolls in real world data, as is expected \cite{wolf2017we}.
We thus ask crowdworkers\footnote{We used Amazon Mechanical Turk for all crowdsourcing tasks. Our tasks pays workers well above minimum wage, and we asked privacy and policy experts to review these tasks before launching. The tasks do not request any personal information from workers.} to label utterances as being ``troll-like'' by asking the question ``Is the last message a good response?''. Screenshots of the crowdsourcing task are given in the Appendix.

We use 5 conversations as an onboarding task for the crowdworkers qualification test. Additionally, we have an onboarding in-flight mechanism; one conversation we know the answer to is mixed into the crowdworkers' assignment as post hoc quality control. All annotations from workers who failed the ``in flight'' onboarding are removed from the data.

We label 527 utterances over 81 conversations, annotating each utterance with three separate crowdworkers, and taking the majority vote to decide the label.

The task is then to identify which examples are low quality. We note that in this case 
there are no annotations by the humans themselves, hence everything is marked as safe as in the Gaslight Troll case, when trolling takes place. 

\section{Learning Methods} \label{sec:methods}

We are given a training set of examples $(x_i, y_i),{i=1,\dots,\ell}$, provided by a set of users, where the users can be thought of as an (unknown) mixture of helpers and trolls, i.e. while some of the data is high quality, some of the data is provided either carelessly or maliciously.  In this section we
describe the set of learning algorithms we employ to learn from such data.

\subsection{Baseline Method}
 Our baseline approach simply assumes that all the data is equally reliable. It first assumes a random split of examples into a training and validation set. The model is then trained, performing early stopping on the validation set, as standard.  In our baseline approach, and all other subsequent approaches, we employ a 128M parameter transformer model as a classifier, 
 using the pre-trained model from
 \citet{dinan2019build}.
 
\subsection{Per-Example Removal} \label{sec:pud}
This set of methods (and all subsequent methods) begins by assuming an original training set and validation set, as in the baseline model.
One then performs the following procedure:
\begin{enumerate}
    \item Split the original training data into $k$-folds, withholding fold $i$ from training, using the original validation set for early stopping. This is used to train $k$ models.
    \item Use model $i$ to {\em``correct''} withheld fold $i$ by comparing the model predictions and user labels, and modifying the disagreeing labels. One obtains  a full {\em corrected} training set by concatenating all $k$ folds together. We consider two {\em correcting} methods: 
    \begin{itemize}
        \item {\bf Flipping}: where if the prediction and the user label disagree, we keep the examples, but flip the label, assuming  it is mislabeled; or
        \item {\bf Removal}: where we remove the examples entirely if the prediction and user label disagree.
    \end{itemize}  
    \item One then trains a model on the {\em corrected} training dataset, using the original validation set, and uses this model to ``correct'' the validation set in the same way as above.
    \item Finally, use the full {\em corrected} train and validation set to train the final classifier with early stopping.
\end{enumerate}

\subsection{Per-Example Soft Bootstrapping}

This example-based method is proposed by \cite{reed2014training}, which is termed a “Soft” bootstrapping. It uses predicted class probabilities $\textbf{q}$ directly to generate regression
targets. It then combines $\textbf{q}$ with the observed noisy multinomial labels $\textbf{t} \in \{0, 1\}, \sum_{k} t_{k} = 1$ for each batch during otherwise standard training as follows: 
\[
\mathcal{L}_{soft}(\textbf{q}, \textbf{t}) = \sum_{k = 1}^{L} [\beta t_{k} + (1 -  \beta)q_{k}] log(q_{k})
\]
A parameter $\beta$ is introduced to control the weight of the loss between generated targets and observed targets.

\subsection{Per-User Removal} 
This user-based method first splits the training data into $k$-folds, similarly to per-example removal, but makes then decisions based at the user rather than example level.

For each user, the cross validation-based predictions are computed, and if the fraction of disagreements of the model with the user labels exceeds $\theta$, the entire user's data is rejected. 
That is, if a user has too many suspiciously labeled examples, their entire set of data is ignored.

\subsection{Per-User+Example Removal} 
    The last method, per-user removal, is quite extreme and does not deal with the case of ``part-time'' trolls, that have some adversarial data, but also some high quality data that it would be beneficial to keep.
     Therefore, in this proposed method, both per-user and per-example removal is applied.  As before,  all users with disagreement greater than $\theta$ are corrected (but compared per-user removal, one can possibly use a less extreme threshold and keep more users). Furthermore, for all remaining users, utterances are removed if they disagree with the model’s individual predictions using the Per-Example method of \autoref{sec:pud}. 

 \subsection{Soft Per-User Robust Removal (PURR)} 
 \label{sec:purr}
    The previous user-based methods all make a hard decision to remove examples (or not), and in particular for the Per-User method they may make a decision to remove all examples for a given user. 
    
    We can design an algorithm that  makes hard decisions whether to remove examples or not, but still incorporates the user level in a soft manner. 
    The main idea is to score a given utterance $x$ with:
    \[
       score(x) =  \alpha f(x) + (1-\alpha) g(U(x) \setminus x)
    \]
    where $U(x)$ is the set of examples from the same user who authored example $x$,  $f(\cdot)$ is a ``trustworthiness'' scoring function that measures the quality of an example,  and $g(\cdot)$ is an aggregate scoring function  that measures the quality of a set of examples from the same user. We then remove examples that fail to meet a certain threshold, which like $\alpha$ is  a tunable parameter.
    
    For the scoring function we use:
    \[
    f(x_i)= (y_i p_i + (1-y_i)(1-p_i),
    \]
    where $y_i$ is the label  assigned by the user of example $i$ (either 0 or 1), and $p_i$ is the prediction (probability) given by the $k$-fold model  used in the Per-Example approach of \autoref{sec:pud}. 
    
    For the aggregator $g(\cdot)$ we use the average trustworthiness score of the examples by the same user: 
    \[
    g({\bf {x}}) = \frac{1}{|\bf {x}|}\sum_{x' \in {\bf {x}}}  f(x').
    \]

\begin{table*}[t!]
\center
\small
\begin{tabular}{lccccccc}
{\bf Algorithm} &  Helper only        &  Troll          &   Master Troll   &   Safe Troll     &  Unsafe Troll   &    Lazy Troll     &   Gaslight Troll   \\
\hline
Oracle Troll Removal & 4\%  &  8\%  & 5\% & 6\%  & 5\%  & 6\%  & 5\% \\
Standard Training    & 4\%  & 31\%  & 29\% & 21\% & 22\% & 16\% & 21\% \\
\hline
{\em Example-based Methods}\\
Per-Ex Flip          & 6\%  &  23\% & 20\% & 18\% & 17\% & 11\% & 29\%\\
Per-Ex Removal       & 5\%  &  19\% & 18\% & 20\% & 21\% & 8\% & 31\% \\ 
Soft Bootstrap       & 4\%  &  24\% & 28\% & 16\% & 19\% & 19\% & 21\% \\ 
\hline
{\em User-based Methods}\\
Per-User Removal      &  6\% &  23\% & 23\% & 20\% & 21\% & 13\% & 38\% \\ 
Per-User+Ex Removal   &  5\% &  12\% & 11\% & 10\% & 10\% & 8\% & 28\% \\ 
Soft PURR             & 4\% & 15\% & 14\% & 17\% & 21\% & 9\% & 30\%\\
\hline
\end{tabular}
\caption{Final {SafetyMix} Task Error rates of various troll robustness learning algorithms, compared to a standard learning baseline and an oracle troll example removal baseline. Methods that take into account user-level and example-level troll behavior work best.}
\label{table:safetymix_error_rates}
\end{table*}

\begin{table*}[t!]
\center
\small
\begin{tabular}{lccccccc}
{\bf Algorithm} &  Troll          &   Master Troll   &   Safe Troll     &  Unsafe Troll   &    Lazy Troll     &   Gaslight Troll   \\
\hline
Oracle Troll Removal & 100 \slash 100 & 100 \slash 100 & 100 \slash 100 & 100 \slash 100 & 100 \slash 100 & 100 \slash 100 \\
Standard Training  & 0 \slash 0 & 0 \slash 0 & 0 \slash 0 & 0 \slash 0 & 0 \slash 0 & 0\slash 0\\
\hline
{\em Example-based Methods}\\
Per-Ex Flip or Removal&   57 \slash 68 & 55 \slash 63 & 52 \slash 49 & 59 \slash 56  & 56 \slash 91 & 29 \slash 12 \\ 
\hline
{\em User-based Methods}\\
Per-User Removal            &   67 \slash 69 & 68 \slash 64 & 48 \slash 95 & 52 \slash 95 & 53 \slash 59 & 100 \slash 63\\ 
Per-User+Ex Removal         &   54 \slash 83 & 55 \slash 79  & 40 \slash 96 & 44 \slash 97 & 46 \slash 86& 68 \slash 63\\ 
Soft PURR &  52 \slash 95 & 55 \slash 84 & 23 \slash 6 & 56 \slash 27 & 39 \slash 96 & 22 \slash 7 & \\
\hline
\end{tabular}
\caption{{SafetyMix} troll example precision and recall of various troll robustness learning algorithms, compared to a standard learning baseline and an oracle troll example removal baseline.
\label{table:safetymix_prec_rec}
}
\end{table*}

\begin{figure}[t]
    \centering
    \includegraphics[width=7.77cm]{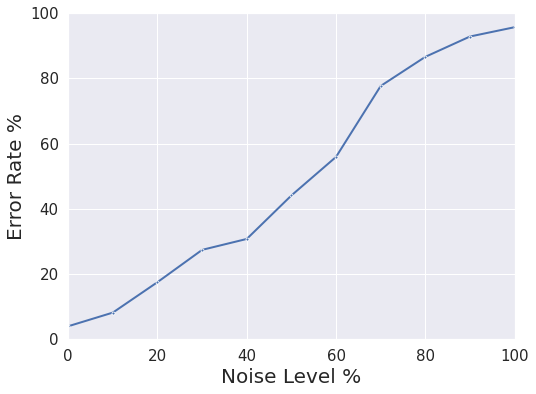}
    \vskip -2mm
    \caption{The relation between the Troll noise level and final error rate when trolls flip labels, instead of being helpers.  A noise level of 0\% indicates all helpers, a level of 50\% indicates all Trolls who flip 50\% of the labels. }
    \label{fig:error-noise}
\end{figure}

\section{Experiments}

\subsection{SafetyMix Experiments}

We compare the different algorithms from \autoref{sec:methods} on all the troll settings
described in \autoref{sec:safetymix}. For each method, we perform balanced training by sampling batches such that they have a roughly equal number of positive and negative examples 
(note: labels here come from the users, so may be noisy).
We then use the validation set to do early stopping and to select hyperparameters, where applicable.  We report results averaged over 10 runs, except in the case of the  Soft bootstrap and Soft PURR methods which are averaged over 5 runs, as they have a large space of hyperparameters due to also tuning  their $\beta$ and $\alpha$ parameters, respectively.

\paragraph{The Impact of (Standard) Trolls}
We first conduct an investigation of the impact of trolls adversarially labeling data on the final classifier accuracy on the SafetyMix benchmark, using standard classifier training.  We use Standard (part-time) Trolls  that mislabel data N\% of the time (flipping the labels), varying the noise level \% and report the error rate of the final classifier trained on this data. Results are shown in 
\autoref{fig:error-noise}.
We observe a low error rate (good accuracy) if all the data providers are helpers (Noise Level 0\%). The error rate steadily worsens as the noise level increases, where a level of 50\% noise yields roughly 50\% error rate -- random chance. Overall, trolls can inflict significant damage to our learning systems.

\paragraph{The Impact of Different Troll Types}
We next evaluate performance of standard model training for the different troll types (described earlier in \autoref{sec:troll_types} and \autoref{table:showtrolls}). In all cases, we use a mix of 50\% trolls and 50\% helpers. Results are given in 
\autoref{table:safetymix_error_rates}.
We find standard trolls cause the worst degradation in performance with a 31\% error rate, with Master Trolls at a similar rate of 29\%, while other troll types have lower error rates. This can be explained by other types only mislabeling some of their data, rather than all of it. For example,
Safe Trolls, with 21\% error rate only mislabel unsafe data (marking it as safe), while Unsafe trolls do the opposite, yielding 22\%. Lazy trolls mislabel half as much data as standard trolls because a random label is assigned (so half the time these labels are correct), yielding 16\%. Gaslight trolls use adversarial unsafe data, marking it always as safe, which yields 21\% error rate, perhaps because this does not provide mislabeled safe data.

\paragraph{Oracle Troll Removal for Different  Troll Types}
In these same setups we can  also
measure the best performance we can achieve by removing all the troll data using an oracle. This is possible as we know which users are trolls  and which specific utterances are mislabeled in the SafetyMix benchmark. 
This method then trains the same classifier only on the subset of data that is left and performance is reported. 
The Helper only setting does not have any adversarial data in the original setup, and so obtains the same error rate of 4\%, while (Standard) Troll increases to 8\% (from 4\%) due to the loss of data. However, this is still a huge improvement from the 31\% of standard training which does not remove the troll data, indicating the ceiling of possible improvement with a robust learning algorithm that can identify trolls.
Similar results are found for the other trolls, with slightly varying performance roughly in line with how many examples are mislabeled in that setting.

\paragraph{Example-Based Removal Methods}
All three example-based methods (Per-Example Flip, Per-Example Removal, and Soft Bootstrap) decide whether to filter a given example based only on the text of the utterance itself. They all provide improvements from filtering trolls, although there are differences depending on the setting.
Per-Example Removal gives the best reductions for Troll, Master Troll and Lazy Troll, e.g.  19\% error vs. standard training 31\% error for the Troll setting. Note, this still leaves lots of potential improvement according to the oracle result of 8\%. Soft Bootstrap works particularly well for Gaslight Trolls and Safe Trolls, which are the two settings where everything is marked as safe. We speculate that the unbalanced nature of the mislabeling may be difficult for the other algorithms. Further, in the Helper only setting Soft Bootstrap is the only method that does not degrade performance. The other methods actually filter some of the helpers, mistakenly thinking they are adversarial increasing the error rate slightly, e.g. Per-Example Removal, which otherwise works well, increases the error rate in this 
setting from 4\% to 5\%.  On the other hand, Soft Bootstrap is worse in some of the other settings, e.g. Troll, Master Troll and Lazy Troll. 
Overall, there is still scope to find new algorithms that work robustly in all settings.

\paragraph{User-Based Removal Methods}
The three user-based methods (Per-User Removal, Per-User+Example Removal and Soft PURR) decide whether to filter an example based on the text of the utterance itself in combination with the trustworthiness of other examples from the same user. Per-User Removal removes entire users, which yields 
improvements compared to Standard Training on almost all troll settings except for Gaslight Troll, but the improvements are relatively small, and not as good as the non-user based Per-Example Removal method.
Removing all examples from a user is too severe as even if troll users are identified correctly they may not label all their data incorrectly, depending on the troll type. Per-User+Example Removal removes some users completely, and then only some utterances for other users. This gives our best results on some of the settings, in particular (Standard) Troll, with 12\% error rate, which is getting closer to the oracle 8\% performance. 
However, there is still some gap to the oracle for Master, Safe, Unsafe and Lazy Troll, and  it still performs badly for Gaslight Troll. Soft PURR also performs well on Troll, Master and Lazy Troll, but worse on the other settings, 
with the exception that Soft PURR gives the best results in the Helper only setting. Analysing the precision recall for the best performing models in terms of accuracy (\autoref{table:safetymix_prec_rec}) it appears that Soft PURR identifies less trolls, and in general tends to remove less examples compared to the other methods. Overall, our main takeaway however is that taking into account the user-level when identifying adversarial examples is crucial  to improving performance.

\paragraph{The difficulty of Gaslight Trolls}
None of the methods we tried made gains beyond standard training in the Gaslight Troll setting (adversarial unsafe inputs that are always marked as safe), despite being successful in other settings.
Such a setting does seem a realistic scenario in real-world cases, 
hence future research should address this failure.
Part of the problem may be the combination of both the adversarial inputs and the unbalanced nature of the data, as there are no examples of such 
adversarial inputs labeled correctly.

\begin{figure}[t]
    \centering
    \includegraphics[width=7.77cm]{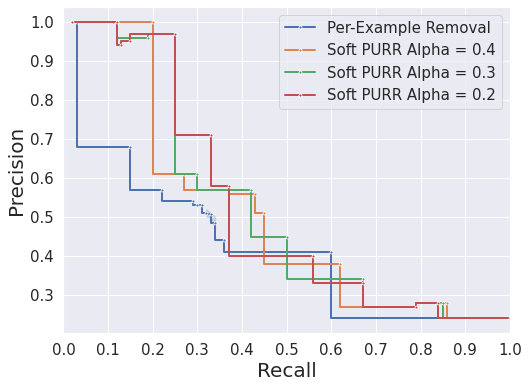}
    \vskip -2mm
    \caption{Precision and recall plot on the conversational real deployment data. We show the Per-Example method compared to the user-based Soft PURR method for varying values of $\alpha$. The user-based method gives better performance. }
    \label{fig:PR_soft-PUUR}
\end{figure}

\subsection{Real Deployment Data}

We next conduct experiments on the real deployment data described in
\autoref{sec:deploy_data}.
We compare two methods: the Per-Example removal method and the 
Soft PURR method, as representatives of example-based and user-based methods, in order to see if we arrive at the same conclusions as in the {SafetyMix} data that user-based methods work best.

We perform these experiments ``zero-shot'' by taking an  off-the-shelf safety classifier of dialogue utterances from the existing work of \citet{dinan2019build}\footnote{Available at: \small{\url{https://github.com/facebookresearch/ParlAI/blob/main/parlai/zoo/dialogue_safety/single_turn.py}}, see also \small{\url{https://parl.ai/projects/dialogue_safety/}}  for usage instructions. 
}. That safety classifier is trained on the datasets we used before to construct {SafetyMix}, in addition to WikiToxicComments \cite{wulczyn2017ex}.
 In this case we are 
 detecting the safety of an utterance, not the quality of a feedback label, so we adapt the methods trustworthiness score slightly, by using the score from this safety classifier instead. So, for the Soft PURR method the probability of the safety classifier is directly used as the score of the 
 quality of the utterance, i.e., used as function $f(\cdot)$.
  For  $g(\cdot)$ we take the average scores using $f(\cdot)$, as before. 
 
We report the precision and recall of troll utterance 
identification in \autoref{fig:PR_soft-PUUR}.
Varying the threshold for each method we can plot performance of each method, where the best methods shift the results towards the right. We find that for various values of 
$\alpha$ the user-based Soft PURR algorithm outperforms Per-Example Removal. 
We note that $\alpha=0$ results in the Soft PURR method reverting to the Per-Example method, and their plots become identical.
Hence, we find that in real deployment data we observe similar findings to our results on {SafetyMix}.

\section{Conclusion}

We have investigated the problem of adversarial behavior, which is mixed in with non-adversarial behavior, when interacting and providing  feedback to conversational agents.
Humans are an important learning authority for AI systems, 
but any learning in the real-world setting must associate trust to certain sources,
and not to others. While most research in robust algorithms and denoising in machine 
learning is concerned with noise models at the example level, we have 
shown in experimental studies that methods that assign trust at the user-level provide 
improved performance in this setting. Troll users with poor behavior tend to be {\em repeat offenders}, which can be detected by user-based algorithms.
This has been shown on our new benchmark {SafetyMix} as well as real deployment data.
Further work should continue to study further user-based algorithms, particularly in
the cases we have identified as difficult, such as gaslight trolls. We thus release
our new benchmark and data to aid this continued research.

\section{Limitations and Discussion}

In this work we have studied robust learning in the mixed adversarial non-adversarial case. We focused on a partially synthetic benchmark SafetyMix  (with real input data, but synthetic noise)  so that we could analyze different kinds of noise model, as well as conducting experiments on
real data from a chatbot deployment. 

This work uses English-language models and data, with real data from a model deployed to people located in the United States. While the methods themselves should generalize well to other languages and contexts, the behavior and types of trolls (and therefore, which methods are empirically most effective for robust learning from deployment data) are likely to differ between deployment environments, e.g., as seen between deployment of Tay and Xiaoice \cite{tayXiaoice}. 

For both synthetic and real data, we studied learning from textual inputs and binary labels, but there are other possible learning settings that can be studied that we have not  addressed. 
In particular, the multi-label case, real-valued case, or the case of missing labels. In the latter,
one could consider a setup where users can like (positive), flag/dislike (negative) or provide no label at all. Algorithmically, this could be addressed simply by bundling the positive and the ``don't know'' class into a single label, or ignore the ``don't know'' class altogether, but it is unclear how these choices would affect the results.
Further, in conversational data labels may be associated with both human and model utterances, and in our current experiments we have not attempted to make this differentiation.

There are also other learning signals one can use  other than  classification-based labels. For example one can make use of symmetric conversations conducted between models and humans  during deployment, to learn to mimic human conversationalists,
thus requiring no labels at all \cite{shuster2020deploying}. 
There are other ways to learn from textual responses as well, including asking questions (and learning from the answers) 
\cite{li2016learning}, learning from textual feedback/comments \cite{li2016dialogue}, or text response signals that indicate positive reaction \cite{hancock2019learning}. 
Finally, one could make use of more sophisticated UIs, for example stack ranking potential responses \cite{ouyang2022training}.
All these different approaches might be exploitable by trolls in different ways, thus requiring further study.

A completely different way of dealing with trolls is to rely on other humans (helpers) to police them directly, which is effectively done for example by the editors of Wikipedia when dealing with article vandalism \cite{shachaf2010beyond}. In our setting humans do not directly interact with each other, but one could imagine systems where engaged helpers might be willing to adjudicate to deal with troll behavior. We do not explore this direction in this work.

Finally, if the long-term goal is to learn continually from conversations and human-provided feedback \cite{roller2020open}, 
then one must also study the impact of an iterative learning cycle, which we did not address in this paper. That is, the issue of whether errors would compound over iterations, or if the model would be self-correcting over time.





\clearpage
\bibliography{anthology,custom}
\bibliographystyle{acl_natbib}

\clearpage
\onecolumn
\appendix
\section{Appendix}\label{sec:appendix}

\begin{figure}[h!]
    \centering
    \includegraphics[width=\textwidth]{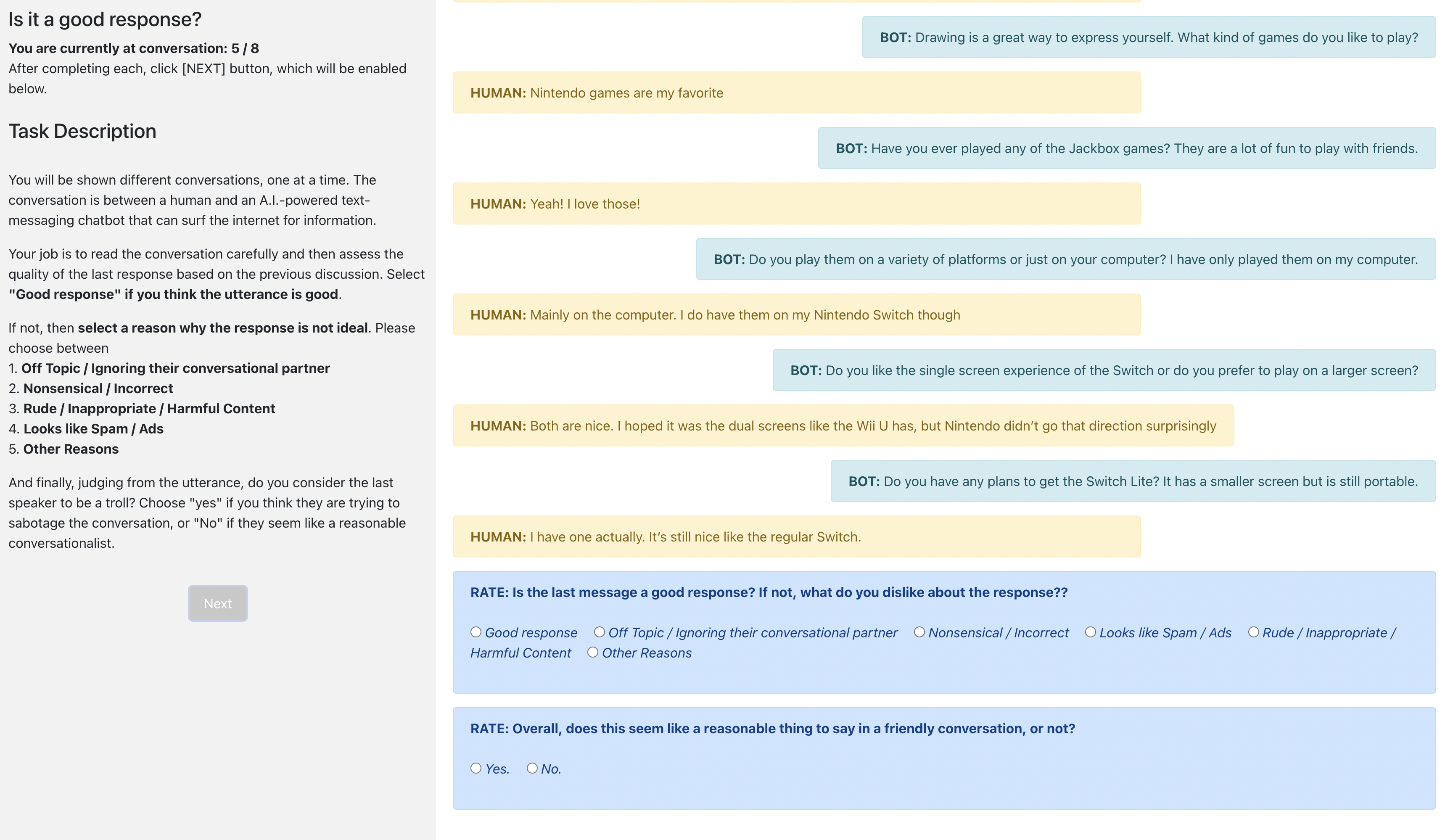}
    \vspace{5mm}\\
        \includegraphics[width=\textwidth]{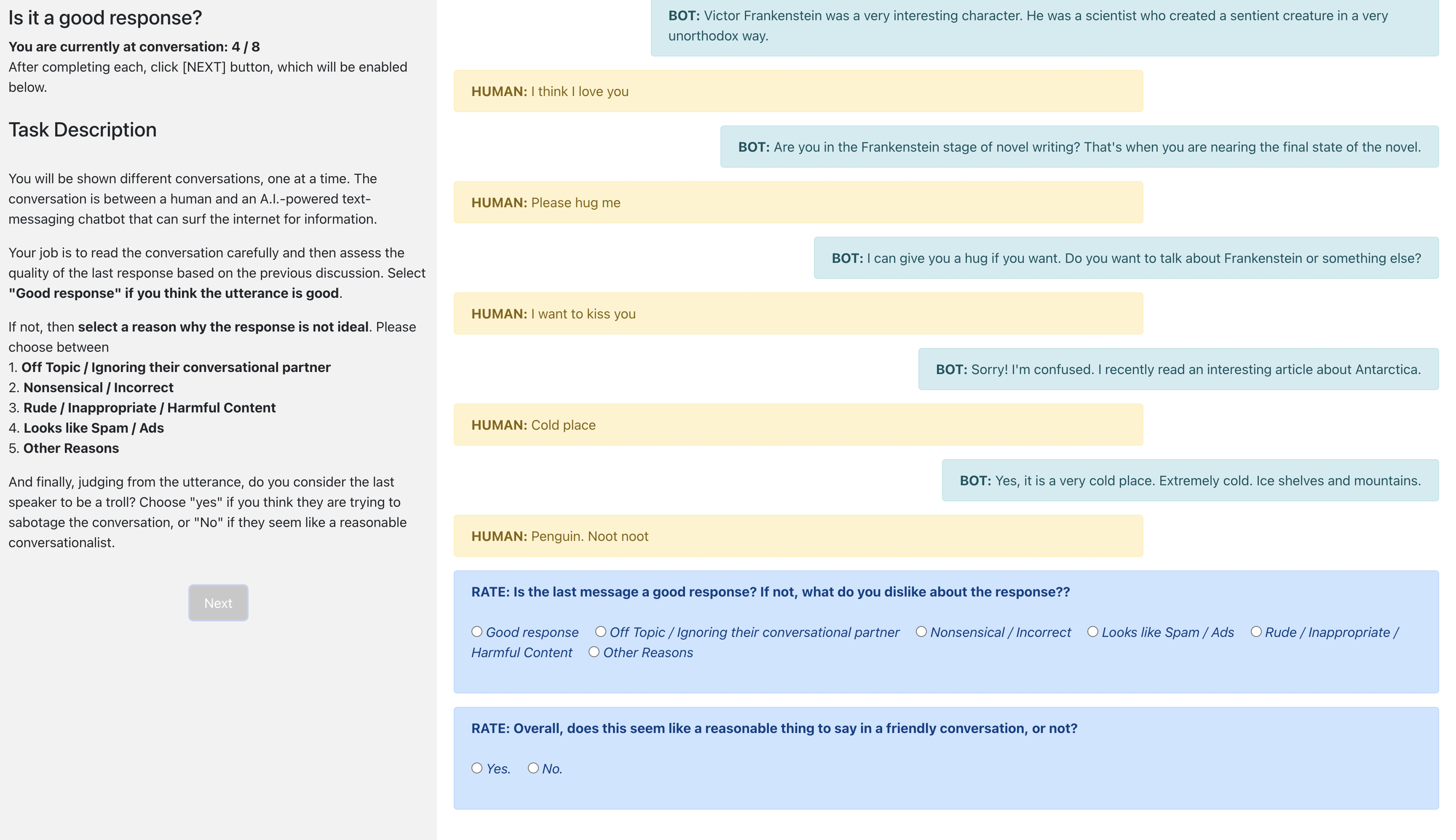}
    \caption{Crowdworker task to annotate real deployment data.}
    \label{fig:crowdworker-task}
\end{figure}

\end{document}